\documentclass[10pt, conference, compsocconf]{IEEEtran}
\ifCLASSINFOpdf
\else
\fi
\usepackage{graphicx}
\usepackage{epstopdf}
\usepackage{algorithm}
\usepackage{amsmath,amsfonts,amssymb}
\usepackage{subcaption}
\usepackage{multirow}
\usepackage{graphicx}
\graphicspath{ {images/} }
\usepackage{paralist}

\begin{document}
%
\title{ATFaceGAN: Single Face Image Restoration and Recognition from Atmospheric Turbulence}


\author{\IEEEauthorblockN{Chun Pong Lau , Hossein Souri and Rama Chellappa}
\IEEEauthorblockA{Center for Automation Research , UMIACS\\
University of Maryland, College Park\\
\tt\small \{cplau,hsouri,rama\}@umiacs.umd.edu}
}


%


\maketitle

\begin{abstract}
Image degradation due to atmospheric turbulence is common while capturing images at long ranges. To mitigate the degradation due to turbulence which includes deformation and blur, we propose a generative single frame restoration algorithm which disentangles the blur and deformation due to turbulence and reconstructs a restored image. The disentanglement is achieved by decomposing the distortion due to turbulence into blur and deformation components using deblur generator and deformation correction generator respectively. Two paths of restoration are implemented to regularize the disentanglement and generate two restored images from one degraded image. A fusion function combines the features of the restored images to reconstruct a sharp image with rich details. Adversarial and perceptual losses are added to reconstruct a sharp image and suppress the artifacts respectively. Extensive experiments demonstrate the effectiveness of the proposed restoration algorithm, which achieves satisfactory performance in face restoration and face recognition. 

\end{abstract}


%
\IEEEpeerreviewmaketitle

\section{Introduction}
Capturing images at long ranges is always challenging as the degradation due to atmospheric turbulence is inevitable. Under the effects of the turbulent flow of air and changes in temperature, density of air particles, humidity and carbon dioxide level, the captured image is blurry and deformed due to variations in the refractive index \cite{hufnagel1964modulation, roggemann1996imaging}. This will significantly degrade the quality of images and performances of many computer vision tasks such as object detection \cite{oreifej2013simultaneous}, recognition and tracking \cite{chen2014detecting}. To suppress these effects, two classical approaches have been considered, one based on adaptive optics \cite{pearson1976atmospheric, tyson2015principles} and the other based on image processing \cite{furhad2016restoring, hirsch2010efficient, zhu2013removing, micheli2014linear, meinhardt2014implementation, lou2013video, lau2019variational, lau2019restoration, chak2018subsampled}. However, these methods require multiple image frames captured by a static imager. Mathematically, \cite{zhu2013removing, hirsch2010efficient, lau2019variational} the process of image degradation due to atmospheric turbulence can be represented as  
\begin{equation} \label{eq: model eq old}
    \Tilde{I}_k = D_k(H_k(I)) + n_k,
\end{equation}
where $\Tilde{I}_k$ is the observed distorted images, $I$ is the latent clear image, $H_k$ is a space-invariant point spread function (PSF), $D_k$ is the deformation operator, which is assumed to deform randomly and $n_k$ is the sensor noise.

Recently, many learning-based face restoration algorithms such as face deblurring \cite{Chrysos_2017_CVPR_Workshops, shen2018deep, lu2019unsupervised} and face superresolution \cite{chen2018fsrnet, Yu_2018_CVPR, Yu_2018_ECCV} have been proposed. Moreover, the emergence of Generative Adversarial Networks (GAN) has further improved the quality of reconstructed images. However, these methods have not tackled the problem of deformation, which greatly reduces the quality of the aquired images and the performance of many computer vision tasks. 

Recently, \cite{chak2018subsampled} proposed a generative method to restore a clean image from multiple frames using a Wasserstein GAN \cite{arjovsky2017wasserstein} and a subsampled frames algorithm proposed by \cite{lau2019variational}. However, the method assumes a multi-frame setting with a static object. This assumption may not be practical in real life situations.

\begin{figure*}[t] 
\centering
\begin{subfigure}[t]{\textwidth}
\includegraphics[width=\textwidth]{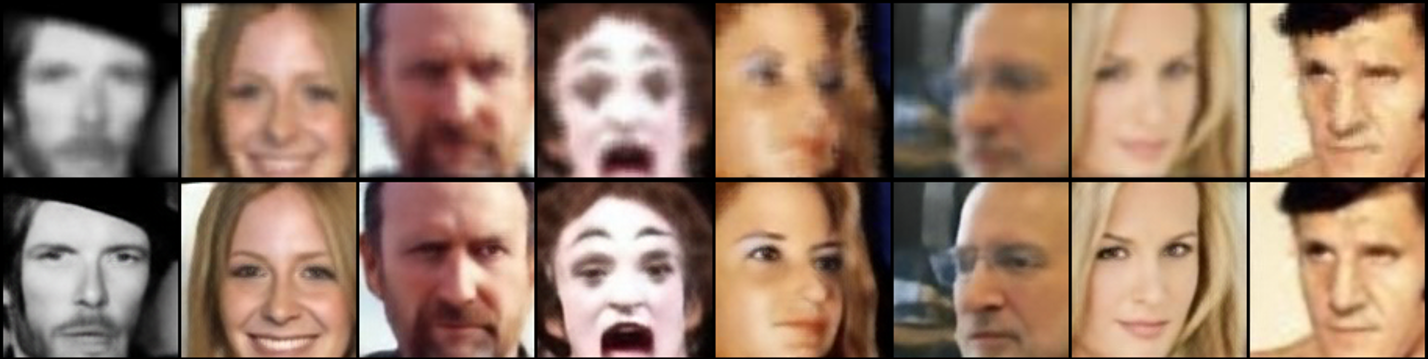}
\vspace{-1em}
\end{subfigure}
\begin{subfigure}[t]{\textwidth}
\includegraphics[width=\textwidth]{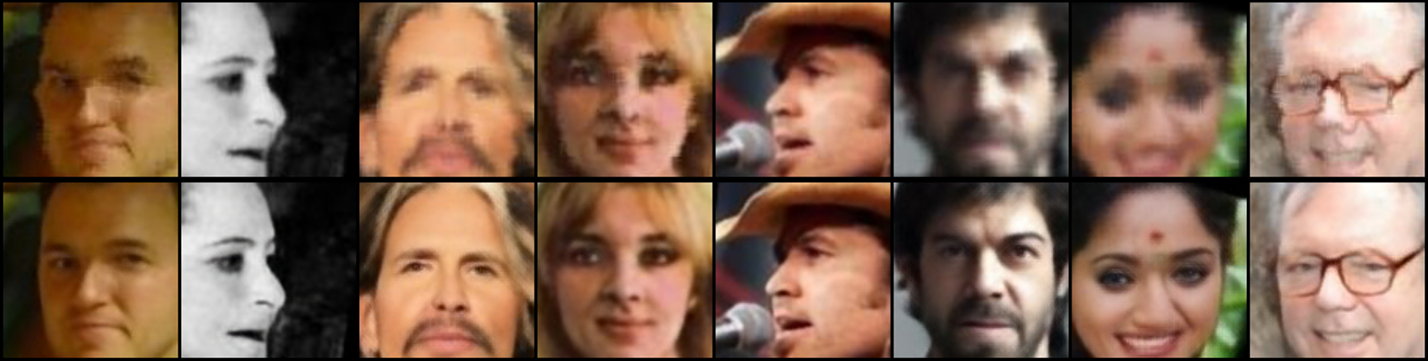}
\vspace{-1em}
\end{subfigure}
\caption{Restored images of ATFaceGAN from the publicly available face dataset \cite{bansal2017umdfaces}. Rows 1 and 3 present the synthetic atmospheric turbulence degraded images. Rows 2 and 4 present the corresponding restored images obtained by the proposed algorithm.}
\label{fig: Restored results of ATFaceGAN}
\end{figure*}

Motivated by the recent success of data-driven approach, we propose a generative single face image restoration algorithm, namely \textbf{A}tmospheric \textbf{T}urbulence \textbf{Face} \textbf{GAN} (\textbf{ATFaceGan}), which reconstructs a clean face image with texture details preserved by simultaneously disentangling blur and deformation. We build two generators, namely, deblur function and deformation correction function to decompose the degradations in turbulence. Also, we propose a two path training approach to further disentangle the degradation and reconstruct two images. A fusion function is used to combine the information in the two restored images and reconstruct a sharp face image. Some sample restored images are shown in Fig. \ref{fig: Restored results of ATFaceGAN}. 

Our contributions are summarized below:
\begin{enumerate}
    \item The proposed method tackles the atmospheric turbulence degradation problem with a single image input.
    \item We propose a generative face restoration algorithm trained in an end-to-end manner, which tackles degradation due to both blur and deformation by building the deblur function and deformation correction function respectively. 
    \item We propose a two path training strategy to further disentangle the blur and deformation and improve the quality of the restored image.
    \item We propose a fusion network to combine the latent features of the intermediate results and reconstruct one clean restored image. 
    \item Experiments demonstrate that the restored face image is satisfactory in both quantitative and visual assessment. Further, the restored face images yields improved recognition performance.
\end{enumerate}
\section{Related Work}
\textbf{Turbulence Degraded Image Restoration} \quad Classical methods of restoring images degraded by turbulence generally include two approaches. One is "lucky imaging" \cite{aubailly2009automated, vorontsov2001anisoplanatic}, which chooses a frame or a number of good frames in a turbulence degraded video and fuses the selected frames. Another one is the registration-fusion approach \cite{hirsch2010efficient, zhu2013removing, xie2016removing, lau2019restoration}, which first constructs a good reference image and aligns the distorted frames with the reference image using a non-rigid image registration algorithm. After alignment, the registered images are fused following which a restoration algorithm is applied to deblur the fused image to obtain the final restored image. Despite having satisfactory results, these methods assume \textit{multi-frame inputs with static objects}. This assumption is violated easily in reality, for example, when pedestrians are moving in long range surveillance videos.  

\textbf{Face Restoration} \quad Due to recent successes of CNNs and GANs, several CNN-based face restoration algorithms have been proposed. \cite{Chrysos_2017_CVPR_Workshops} proposed a CNN with Residual Blocks to deblur face images. \cite{shen2018deep} proposed a multi-scale CNN that exploits global semantic priors and local structural constraints for face image deblurring as a generator and built a discriminator based on DCGAN \cite{radford2015unsupervised}. \cite{lu2019unsupervised} proposed an unsupervised method for domain-specific single image deblurring by disentangling the content information and blur information using the KL divergence constraint and improves the performance of face recognition. However, since \textit{degradation due to turbulence contains motion blur, out-of-focus blur or compression artifacts}, these methods could not obtain satisfactory results. 

\begin{figure*}[t] 
\centering

\includegraphics[width=\textwidth]{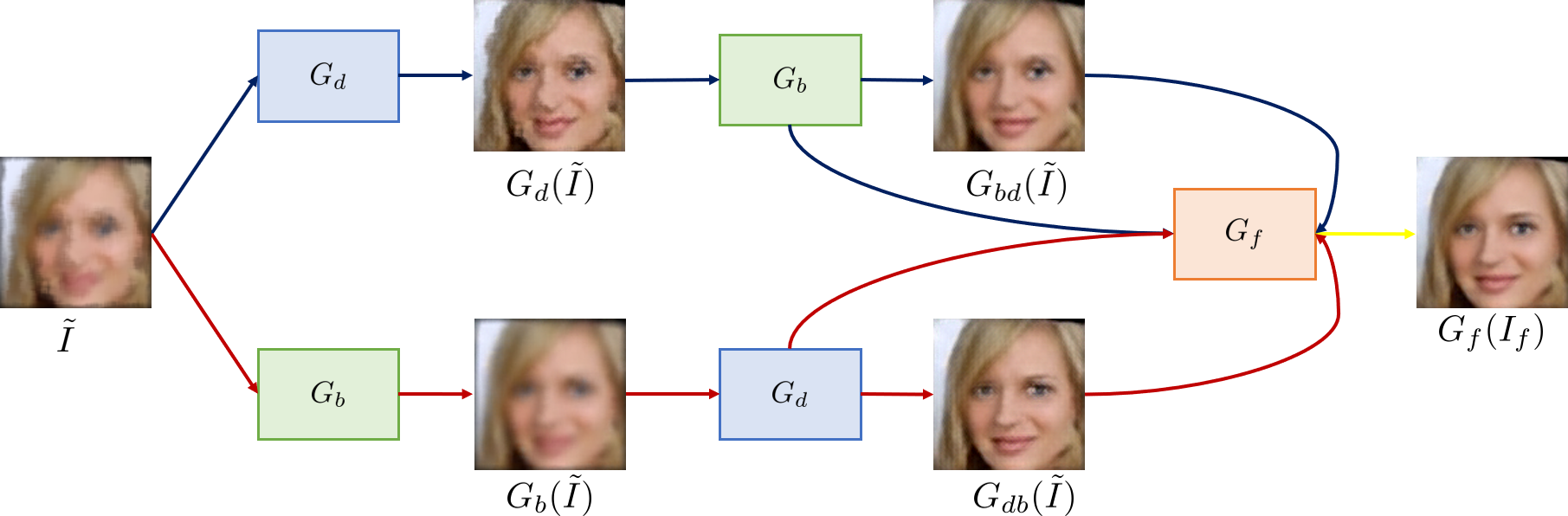}

\caption{Overview of the proposed restoration algorithm. Given a distorted image $\Tilde{I}$, it is first passed through the deformation correction function $G_d$ to get it aligned and deblurred with deblur function $G_b$ and generate $G_{bd}(\Tilde{I})$. It is then passed through the bottom branch (red), gets deblurred by $G_d$ and then algined by $G_b$ to obtain $G_{db}(\Tilde{I})$. Features of $G_{bd}(\Tilde{I})$ and $G_{db}(\Tilde{I})$ are concatenated and passed through the fusion function $G_f$ to obtain the final restored image $G_f(I_f)$.}
\label{fig: Overview}
\end{figure*}

\section{Proposed Algorithm}
The proposed face image restoration algorithm is trainable in an end-to-end manner. Our goal is to reconstruct a sharp face image from the distorted face image and enhance the performance of face recognition systems.

\subsection{Problem Setting}
Following the formulation of the degradation model discussed in \cite{lau2019restoration, zhu2013removing, lau2019variational}, we assure the mathematical model in (\ref{eq: model eq old}). This is the general setting for restoring the latent clean image from a sequence of turbulence-degraded image frames. However, we assume only one frame is available to reconstruct the latent clean image, a more challenging and practical problem than considered earlier. As a result, the subscript $k$ is removed. Also, we notice that the "mixing" of deformation and blur in realistic turbulence face images is very fast and we could not be sure whether deformation precedes blur or blur precedes deformation. Therefore, we use a general turbulence function $T$ to replace $D \circ G$ in (\ref{eq: model eq old}).  Hence, our model becomes 
\begin{equation} \label{eq: model eq}
    \Tilde{I} = T(I) + n.
\end{equation}
Let $\mathcal{I}_b = \{ I_b := H(I) | I \in \mathcal{I}\}$ and $\mathcal{I}_d= \{ I_d := D(I) | I \in \mathcal{I}\}$ be the space of blurry and deformed face images respectively. Our goal is to construct a restoration function $G$ to restore the distorted face images, i.e. $G(\Tilde{I}) = I$. However, it is a highly ill-posed problem as we have very little prior information to reconstruct $\Tilde{I}$. Hence, a data-driven approach, in particular the Wasserstein GAN with gradient penalty, is applied to restore it. Moreover, blur and deformation are always combined in the turbulence-degraded face images. We hope to build a deblur function $G_d$ and a deformation correction function $G_b$ to remove the undesired blur and deformation, i.e. $G_d(\Tilde{I}) = I_d$ and $G_b(\Tilde{I}) = I_b$. Therefore, we split the turbulence degradation due to blur and deformation in the training stage. In order to restore a general turbulence function $T$ which contains both the blurring operator $H$ and the deformation operator $D$, we propose a two path training approach, which tries to obtain more information to obtain a better result. Therefore, two restored images are obtained, i.e. $G_d(G_b(\Tilde{I}))$ and $G_b(G_d(\Tilde{I}))$. A fusion network is implemented to improve the restoration results. Denote $G_{bd} = G_b \circ G_d$, $G_{db} = G_d \circ G_b$ and $F(I)$ be the features of image $I$. Mathematically, 
\begin{equation}
    G(\Tilde{I}) = G_f(I_f),
\end{equation}
where $G_f$ is a image fusion function and $I_f$ is the feature pairs $(F(G_{db}(\Tilde{I})), F(G_{bd}(\Tilde{I}))$. The end-to-end architecture is illustrated in Fig. \ref{fig: Overview}.

\subsection{Data Augmentation} \label{sec: Data augmentation}
In order to apply a data-driven method to restore a clean face image from distorted faces, sufficient amount of  synthetic training data are needed. Therefore, the blur operator $H$ and the deformation operator $D$ are required to synthesize the distorted images. In this paper, we use the turbulence generation algorithm from \cite{lau2019variational, lau2019restoration, chak2018subsampled} due to its efficiency in choosing different parameters to generate turbulence-degraded images with various severity. 

We follow the procedure discussed in \cite{ lau2019variational, lau2019restoration, chak2018subsampled} to generate a random motion vector field to deform the face images. $M$ points are selected in a face image $I$. For each point $(x, y)$, a $N \times N$ patch $P^N_{x,y}$ centered at $(x, y)$ is considered. A random motion vector field $V_{x,y}$ is obtained in $P^N_{x,y}$. Mathematically, 
\begin{equation}
    V_{x,y} = \eta(G_\sigma \ast \mathcal{N}_1, G_\sigma \ast \mathcal{N}_2),
\end{equation}
where $G_\sigma$ is the Gaussian kernel with standard deviation $\sigma$, $\eta$ is the strength value, $\mathcal{N}_1$ and $\mathcal{N}_2$ are randomly selected from a Gaussian distribution. The overall motion vector field is generated after $M$ iterations as follows:
\begin{equation}
    V = \sum_{i=1}^M V_{(x,y)_i}
\end{equation}
Then this motion vector field would be our deformation operator $D$ defined as 
\begin{equation}
    D(I) = I \boxplus V,
\end{equation}
\noindent where $\boxplus$ is the warping operator. The blurring operator $H$ is simply a Gaussian kernel. For more details, please see \cite{ lau2019variational, lau2019restoration, chak2018subsampled}.

In order to construct the deblur function $G_d$ and the deformation correction function $G_b$, we need to generate a blurry image $I_b$, a deformed image $I_d$ and a distorted images $\Tilde{I}$ from each clean face image $I$. To generate $I_b$, Gaussian blurring filter with parameter $\tau$ is applied on $I$ to get $I_b$. To obtain $I_d$, the random motion vector field with strength $\eta$ is applied on $I$. 

\subsection{Network Architecture}
A Wasserstein GAN with gradient penalty is applied to restore the distorted face images. The generator architecture is a CNN, similar to \cite{kupyn2018deblurgan} used for image deblurring. It contains two strided convolution blocks with stride $12$, six residual blocks \cite{he2016deep} (ResBlocks) and two transposed convolution blocks. There are one convolution layer, instance normalization layer \cite{ulyanov2016instance}, ReLU activation \cite{nair2010rectified} and a Dropout layer with $p=0.5$ in each ResBlock. A global skip connection mentioned in \cite{kupyn2018deblurgan} is also added. The deblur function $G_d$ and deformation correction function $G_b$ are included in both paths of this architecture. The fusion network $G_f$ takes the concatenation of the features from face images $G_{bd}\Tilde{I})$ and $G_{db}(\Tilde{I})$ as inputs. The features are extracted after the activation function of the third ResBlock in $G_b$ and $G_d$. The architecture of the fusion network $G_f$ is exactly the latter half of the structure of $G_b$ and $G_d$, which contains three ResBlock and two transposed convolution blocks. Since the input is the concatenation of two feature vectors, the number of channels in the ResBlocks of $G_f$ is doubled. In order to keep the global skip connection, which has been shown to converge faster, pixel-wise average of $G_{bd}\Tilde{I})$ and $G_{db}(\Tilde{I})$ is added to $G_f(I_f)$. During training, three discriminators, namely $D_b$, $D_d$ and $D_f$, are designed. $D_b$, $D_d$ and $D_f$ determine whether $G_b(\Tilde{I})$ and $G_d(\Tilde{I})$ and $G_f(\Tilde{I})$ are real or fake. The discriminators are Wasserstein GAN \cite{arjovsky2017wasserstein} with gradient penalty \cite{gulrajani2017improved} (WGAN-GP). Their architectures are same as PatchGAN \cite{isola2017image, li2016precomputed}. All the convolutional layers except the last are followed by InstanceNorm layer and LeakyReLU \cite{xu2015empirical}.

\subsection{Disentanglement of Blur and Deformation}
In order to disentangle the turbulence distortion into blur and deformation, the deblur function $G_d$ and the deformation correction function $G_b$ are built. The content loss $\mathcal{L}_{con}$ is defined as 
\begin{equation}
    \mathcal{L}_{con} = \| G_b(\Tilde{I}) - I_b \|_1 + \| G_d(\Tilde{I}) - I_d \|_1,
\end{equation}
which is the sum of the $L_1$ loss between aligned image $G_b(\Tilde{I})$ and $I_b$ and the $L_1$ loss between deblurred image $G_d(\Tilde{I})$ and $I_d$.

\subsection{Two path training}
The two path training strategy helps to disentangle the blur and deformation effects. One fixed order of restoration is needed if two path training is not implemented. For example, the distorted image is restored by $G_b$ and followed by $G_d$ according to (\ref{eq: model eq}). Then during the training phase, $G_b$ is trained with the turbulence degraded images which are both blurry and deformed. In other words, the training images for $G_b$ are implicitly assumed to be both blurry and deformed but not merely deformed. Therefore, if two path training is used, then $G_b$ could learn from turbulence degraded images $\Tilde{I}$ and the deblurred images $G_d(\Tilde{I})$.  

Moreover, the search space of the optimization problem is larger because no implicit structure of degradation is assumed. As the turbulence function $T$ only consists of blur and deformation but not the order of degradation, this gives more information ($G_{bd}\Tilde{I})$ and $G_{db}(\Tilde{I})$) to the network and improve the performance.

\subsection{Fusion Loss}
After both restored images $G_{bd}(\Tilde{I})$ and $G_{db}(\Tilde{I})$ are obtained, their features are fused together to obtain the final restored image. The fusion loss is defined as the $L_1$ loss of the restored image and the real clean image $I$, i.e.
\begin{equation}
    \mathcal{L}_{f} = \| G_f(I_f) - I \|_1. 
\end{equation}

\subsection{Adversarial Loss}
The Wasserstein-1 distance in WGAN has been shown to have good convergence property and is more stable in training given that the function is 1-Lipschitz. To enforce the 1-Lipschitz constraint, gradient penalty is applied. Then the discriminator and generator losses are defined as 
\begin{equation}
\begin{split}
    \mathcal{L}_{Dis}^{\mathcal{I}_i} &= \mathbb{E}_{\Tilde{I} \sim \mathcal{\Tilde{I}}} [D_i(G_i(\Tilde{I}))] - \mathbb{E}_{I_i \sim \mathcal{I}_i} [D_i(I_i))] \\ 
    &+ \lambda_{WGAN} \cdot \mathbb{E}_{\hat{I}_i \sim \widehat{\mathcal{I}}_i} [(\| \nabla_{\hat{I}_i} D_i(\hat{I}) \|_2 -1)^2],
\end{split}
\end{equation}
\begin{equation}
\begin{split}
    \mathcal{L}_{Dis}^{\mathcal{I}_f} &= \mathbb{E}_{I_f \sim \mathcal{I}_f} [D_f(G_f(I_f))] - \mathbb{E}_{I \sim \mathcal{I}} [D_f(I))] \\ 
    &+ \lambda_{WGAN} \cdot \mathbb{E}_{\hat{I}_f \sim \widehat{\mathcal{I}_f}} [(\| \nabla_{\hat{I}} D_f(\hat{I}) \|_2 -1)^2],
\end{split}
\end{equation}
\begin{equation}
    \mathcal{L}_{Gen}^{\mathcal{I}_i} = -\mathbb{E}_{I_i \sim \mathcal{I}_i} [D_i(G_i(\Tilde{I}))]
\end{equation}
\begin{equation}
    \mathcal{L}_{Gen}^{\mathcal{I}_f} = -\mathbb{E}_{I_f \sim \mathcal{I}_f} [D_f(G_f(I_f))]
\end{equation}
where $\widehat{\mathcal{I}}_i$ is the distribution obtained by randomly interpolating between real images $I_i$ and restored images $G_i(\Tilde{I})$, $i \in \{b,d\}$ and $j \in \{b,d,f\}$.  
The adversarial loss is 
\begin{equation}
    \mathcal{L}_{adv} = \sum_{j \in \{ b,d,f \}} \mathcal{L}_{Dis}^{\mathcal{I}_j} + \mathcal{L}_{Gen}^{\mathcal{I}_j}
\end{equation}

\subsection{Perceptual Loss}
Using $L_2$ loss or $L_1$ loss merely as the content loss would lead to blurry artifacts and loss in texture details as these losses favor pixelwise averaging. On the other hand, Perceptual Loss, which is an $L_2$ loss function between the feature maps of real image and generated image, has been demonstrated to be beneficial for image restoration tasks \cite{shen2018deep, kupyn2018deblurgan, lu2019unsupervised}. Therefore, perceptual loss is adopted, which includes
\begin{align}
    \mathcal{L}_p^{\mathcal{I}_i} &=  \| \phi_l(G_i(\Tilde{I})) - \phi_l(I_i)\|_2^2, \quad i \in \{ b,d \} \\
    \mathcal{L}_p^{\mathcal{I}_f} &=  \| \phi_l(G_f(I_f)) - \phi_l(I)\|_2^2 
\end{align}
where $\phi_l(\cdot)$ is the features of the $l^{\text{th}}$ layer of a pretrained CNN. In this paper, the $\texttt{conv}_{3,3}$ layer of VGG-19 \cite{simonyan2014very} network pretrained on ImageNet \cite{deng2009imagenet} is adopted. The total perceptual loss is 
\begin{equation}
    \mathcal{L}_p =  \sum_{i \in \{ b,d \}} \mathcal{L}_p^{\mathcal{I}_i} + \mathcal{L}_p^{\mathcal{I}_f} 
    = \mathcal{L}_p^{\mathcal{I}_b, \mathcal{I}_d} + \mathcal{L}_p^{\mathcal{I}_f} 
\end{equation}
The full loss function is a weighted sum of all the losses,
\begin{equation}
    \mathcal{L} = \mathcal{L}_{adv} + \lambda_{con} \mathcal{L}_{con} + \lambda_{f} \mathcal{L}_{f} + \lambda_{p}^{b,d} \mathcal{L}_p^{\mathcal{I}_b, \mathcal{I}_d} + \lambda_{p}^f \mathcal{L}_{p}^{\mathcal{I}_f} 
\end{equation}
The weights are empirically set for each loss to balance their importance. 
\subsection{Testing}
At test time, only the generators are used. Given a turbulence distorted image $\Tilde{I}$, the restored image is generated as follows:
\begin{equation}
    I^r = G_f(F(G_{bd}(\Tilde{I})), F(G_{db}(\Tilde{I}))).
\end{equation}

\begin{table*}[t!]
\centering

\caption{Ablation study tested with LFW dataset}
\label{tab:ablation}
\begin{tabular}{|l || c | c | c | c | c |}
\hline
  
\multirow{2}{*}{Method} & \multirow{2}{*}{Degraded images} & \multirow{2}{*}{One generator} & {Decompose into } & {Add two path} & {Add fusion } \\ 
& & & two generators & training & function \\ \hline
PSNR & 24.17 & 25.99 & 25.90 & 26.16 & \textbf{27.29}  \\ \hline
SSIM & 0.878 & 0.901 & 0.897 & 0.902 & \textbf{0.924}  \\ \hline

\end{tabular}
\end{table*}

\begin{figure}[t] 
\centering
\begin{subfigure}[t]{0.07\textwidth}
\includegraphics[width=\textwidth]{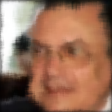}
\caption{}
\end{subfigure}
\begin{subfigure}[t]{0.07\textwidth}
\includegraphics[width=\textwidth]{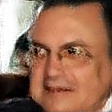}
\caption{}
\end{subfigure}
\begin{subfigure}[t]{0.07\textwidth}
\includegraphics[width=\textwidth]{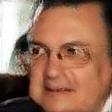}
\caption{}
\end{subfigure}
\begin{subfigure}[t]{0.07\textwidth}
\includegraphics[width=\textwidth]{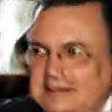}
\caption{}
\end{subfigure}
\begin{subfigure}[t]{0.07\textwidth}
\includegraphics[width=\textwidth]{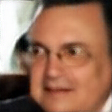}
\caption{}
\end{subfigure}
\begin{subfigure}[t]{0.07\textwidth}
\includegraphics[width=\textwidth]{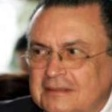}
\caption{}
\end{subfigure}
\caption{Ablation study. (a) is the distorted image and (f) is the sharp image. (b) only contains one generator. (c) is split into $G_d$ and $G_b$. (d) adds two path training and (e) adds fusion network}
\label{fig: Ablation study}
\end{figure}

\section{Experiments}
Our algorithm is trained on \cite{bansal2017umdfaces} and evaluated on six face recognition datasets, including LFW \cite{huang2008labeled}, CFP \cite{sengupta2016frontal}, AgeDB \cite{moschoglou2017agedb}, CALFW \cite{zheng2017cross}, CPLFW \cite{zheng2018cross} and VGGFace2 \cite{cao2018vggface2}.

\subsection{Training details}
The end-to-end design is implemented in Pytorch \cite{paszke2017automatic}. The training was performed on two GeForce RTX 2080 Ti GPU. In training, 10000 aligned face images are randomly picked, which are with resolution $112 \times 112$ from \cite{bansal2017umdfaces} with the turbulence degradation algorithm in Sec(\ref{sec: Data augmentation}) and a batch size of $16$. During training, we use the Adam solver \cite{kingma2014adam} with hyper-parameters $\beta_1=0.9, \beta_2=0.999$ to perform five steps of update on discriminators and then one step on generators. The learning rate is initially set at 0.0001 for the first 30 epochs, then linear decay is applied for the next 20 epochs. For hyper-parameters in deformation operator $D$, we empirically set $\eta = 0.13, N=4, \sigma = 16$ and $M=[1000, 3000, 7000, 10000]$. For hyper-parameters in blurring operator $H$, the parameter $\mu$ is set to be $[1,2,3,4]$. For hyper-parameters in the loss function, we empirically set $\lambda_{con}  = \lambda_{f} = 1000$, $\lambda_{p}^{b,d} = 10$ and $\lambda_{p}^f = 1$. Note that various parameters in $M$ and $\tau$ are randomly picked to synthesize various strength of blur and deformation. The computation time of restoring a $112 \times 112$ image is $0.031$ seconds per image on average. 

\subsection{Testing details}
In all the six testing dataset, all the pairs of the face images are degraded by the algorithm from  \cite{lau2019variational}. PSNR and SSIM are used for evaluating the quality of the restored image. We use a pretrained face recognition network \cite{xu2017high}, which is trained as reported in \cite{guo2016ms}, to test the face verification performance\footnote{Please refer to the corresponding project page for the face verification policy: https://github.com/ZhaoJ9014/face.evoLVe.PyTorch}.

\begin{figure*}[t] 
\centering
\begin{subfigure}[t]{0.115\textwidth}
\includegraphics[width=\textwidth]{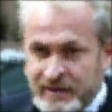}
\end{subfigure}
\begin{subfigure}[t]{0.115\textwidth}
\includegraphics[width=\textwidth]{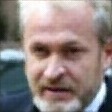}
\end{subfigure}
\begin{subfigure}[t]{0.115\textwidth}
\includegraphics[width=\textwidth]{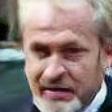}
\end{subfigure}
\begin{subfigure}[t]{0.115\textwidth}
\includegraphics[width=\textwidth]{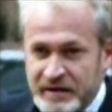}
\end{subfigure}
\begin{subfigure}[t]{0.115\textwidth}
\includegraphics[width=\textwidth]{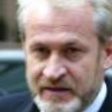}
\end{subfigure}
\\
\begin{subfigure}[t]{0.115\textwidth}
\includegraphics[width=\textwidth]{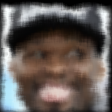}
\end{subfigure}
\begin{subfigure}[t]{0.115\textwidth}
\includegraphics[width=\textwidth]{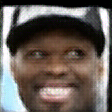}
\end{subfigure}
\begin{subfigure}[t]{0.115\textwidth}
\includegraphics[width=\textwidth]{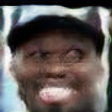}
\end{subfigure}
\begin{subfigure}[t]{0.115\textwidth}
\includegraphics[width=\textwidth]{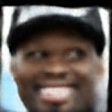}
\end{subfigure}
\begin{subfigure}[t]{0.115\textwidth}
\includegraphics[width=\textwidth]{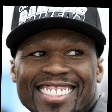}
\end{subfigure}
\\
\begin{subfigure}[t]{0.115\textwidth}
\includegraphics[width=\textwidth]{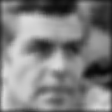}
\caption{}
\end{subfigure}
\begin{subfigure}[t]{0.115\textwidth}
\includegraphics[width=\textwidth]{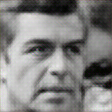}
\caption{}
\end{subfigure}
\begin{subfigure}[t]{0.115\textwidth}
\includegraphics[width=\textwidth]{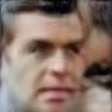}
\caption{}
\end{subfigure}
\begin{subfigure}[t]{0.115\textwidth}
\includegraphics[width=\textwidth]{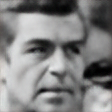}
\caption{}
\end{subfigure}
\begin{subfigure}[t]{0.115\textwidth}
\includegraphics[width=\textwidth]{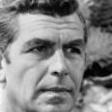}
\caption{}
\end{subfigure}

\caption{Visual performance comparison with state-of-the-art methods. (a) is the distorted image. (b) \cite{kupyn2018deblurgan}. (c) \cite{lu2019unsupervised}. (d) Ours. (e) Groundtruth.}
\label{fig: comparsion}
\end{figure*}

\begin{table*}[t!]
\centering
\caption{Quantitative performance comparison with state-of-the-art
methods}
\label{tab:comparsion psnr ssim}
\begin{tabular}{|c||c | c | c | c | c | c|c|c|c|c|c|c|c|}
\hline
\multirow{2}{*}{Dataset}&Sharp&\multicolumn{3}{c|}{Distorted} & \multicolumn{3}{c|}{\cite{kupyn2018deblurgan}} &  \multicolumn{3}{c|}{\cite{lu2019unsupervised}} & \multicolumn{3}{c|}{Ours}\\\cline{2-14}
  &  Acc & PSNR & SSIM & Acc  & PSNR & SSIM & Acc & PSNR & SSIM & Acc & PSNR & SSIM & Acc \\ \hline 
AgeDB \cite{moschoglou2017agedb} & 0.981  &   22.71 & 0.769 & 0.819 & 24.04 & 0.781 & 0.830 &  21.22& 0.657& 0.750 & \textbf{25.24} & \textbf{0.838} & \textbf{0.835} \\ \hline  
CALFW \cite{zheng2017cross} & 0.959 &    22.85 & 0.831 & 0.842 & 24.43 & 0.843 & 0.844  &21.97 &0.771 & 0.780 & \textbf{25.78} & \textbf{0.890} & \textbf{0.857} \\ \hline
CFP$\_$FF \cite{sengupta2016frontal} & 0.997 & 22.13 & 0.830 & 0.892 & 22.88 & 0.833 & 0.916  &21.40 &0.683 & 0.861 & \textbf{24.37} & \textbf{0.889} & \textbf{0.922}  \\ \hline
CFP$\_$FP \cite{sengupta2016frontal} & 0.981 &  22.94 & 0.850 & 0.799 & 23.84 & 0.850 & 0.812  &21.83 &0.629 & 0.743 & \textbf{25.20} & \textbf{0.901} & \textbf{0.815} \\ \hline
CPLFW \cite{zheng2018cross} & 0.925 &    24.21 & 0.875 & 0.787 & 26.11 & 0.882 & 0.797  &22.68 &0.787 & 0.732 & \textbf{27.29} & \textbf{0.919} & \textbf{0.800} \\ \hline
LFW  \cite{huang2008labeled} &0.998 &    24.17 & 0.878 & 0.936 & 25.80 & 0.884 &\textbf{0.951} &23.10 &0.824 & 0.896 & \textbf{27.29} & \textbf{0.924} & 0.946 \\ \hline
VggFace2 \cite{cao2018vggface2} & 0.952 &    23.44 & 0.849 & 0.837 & 24.99 & 0.856 & 0.853  & 22.06&0.774 & 0.784 & \textbf{26.16} & \textbf{0.896} & \textbf{0.854} \\ \hline

\end{tabular}
\end{table*}

\subsection{Ablation study} \label{sec: Ablation study}
In this section, the results of an ablation study preformed to analyze the effectiveness of each component or loss in the proposed algorithm are presented. Both quantitative and qualitative results on face dataset in \cite{bansal2017umdfaces} are evaluated for the following four variants of our methods where each component is gradually added: 1) only one generator and one discriminator; 2) splitting the generators into two, $G_b$ and $G_d$, and the restored image is $G_b(G_d(\Tilde{I}))$; 3) Applying two path training and the restored image is $G_b(G_d(\Tilde{I}))$ and 4) adding fusion network and fuse them by $G_f$ 

We present the PSNR and SSIM for each variant in Table \ref{tab:ablation} and visual comparisons in Fig. \ref{fig: Ablation study}. From Fig. \ref{fig: Ablation study}, we observe that the resultant images with direct restoration, which only uses one generator, is not satisfactory. This is because turbulence degradation is a very ill-posed problem. There is a large gap between turbulence-degraded and clean image and one generator could not provide enough information to the network. By decomposing the network into two generators, the quantitative performance is similar to one generator but it is less noisy. This is because we have more information for the generators to learn as the intermediate results ($G_d(\Tilde{I})$) provides additional supervision to the final restored image. When we apply the two path training step and as both $I_b$ and $I_d$ are added to supervise the training, the results are good even groundtruth $I$ is not used in the training. Adding the fusion network further improves the result as more information (features of $G_{bd}(\Tilde{I})$ and $G_{db}(\Tilde{I})$) is given to the network and the information is combined by the fusion function $G_f$. Table \ref{tab:ablation} also justifies the result. 

\subsection{Qualitative and quantitative Evaluation} \label{sec: Qualitative and quantitative Evaluation}
Since the proposed algorithm is the first single frame-based image restoration method with turbulence-degraded images, which involve blur and deformation, it is hard to compare with other methods. Therefore, we compare with some state-of-the-art image restoration methods including \cite{kupyn2018deblurgan, lu2019unsupervised}, which could train with our turbulecnce-dagraded image dataset. These two methods are the representative methods for applying GAN in deblurring in supervised and unsupervised ways respectively. For \cite{kupyn2018deblurgan}, we change the batch size from $1$ to $16$ and the number of training epoch to $100$. For \cite{lu2019unsupervised}, we use the default setting.   

The quantitative results are shown in Table \ref{tab:comparsion psnr ssim} and the visual comparison are illustrated in Fig. \ref{fig: comparsion}. In Fig. \ref{fig: comparsion}, we have demonstrated three images: one from LFW, one from CFF and one from AGEDB. The top one is a frontal image with mild blur and mild deformation, the middle one is a frontal image with moderate blur and severe deformation and the bottom one is a non-frontal gray-scale face image with severe blur and mild deformation. For the top image, we can see that blur is suppressed in all three methods. \cite{kupyn2018deblurgan} and \cite{lu2019unsupervised} shows sharper visual result then ours. However, the result from \cite{kupyn2018deblurgan} is noisy and that from \cite{lu2019unsupervised} is deformed. The proposed method restores the image effectively. On the other hand, if both blur and deformation exist, \cite{kupyn2018deblurgan} would induce more noise as shown in Fig. \ref{fig: comparsion} (b) and \cite{lu2019unsupervised} could not remove the deformation as shown in Fig. \ref{fig: comparsion} (c). The proposed method suppresses both blur and deformation. Moreover, as our training set only consists of $10000$ images, which include both colored and grey-scale images, the quantitative results generated by \cite{lu2019unsupervised} are not good compared to \cite{kupyn2018deblurgan} and the proposed method as the number of training sample is not large enough. The proposed method trained with a relatively small training set is effective in the presence of severe blur, deformation and pose. The PSNR and SSIM in Table \ref{tab:comparsion psnr ssim} both demonstrate that the proposed method performs better than state-of-the-art methods. 

For the face verification task, we note that \cite{kupyn2018deblurgan} is slightly better than the proposed method in one out of seven experiments even though both the visual quality and quantitative results of the proposed method is better than \cite{kupyn2018deblurgan}. Except LFW, the proposed method is more accurate than the other two methods. The verification accuracy of \cite{kupyn2018deblurgan} is comparable with the proposed method. It is because \cite{kupyn2018deblurgan} uses only perceptual loss as their content loss. As a result, the restored image from \cite{kupyn2018deblurgan} is perceptually similar than the proposed method. Using the $L_2$ distance from two feature output from $\texttt{conv}_{3,3}$ layer of VGG-19 \cite{simonyan2014very} network as a perceptual metric, namely $d_{VGG}$, we found that the $d_{VGG}$ between restored image by \cite{kupyn2018deblurgan} and the original clean image is $110.82$ in LFW  while the $d_{VGG}$ between the restored image by the proposed method and original clean image is $118.55$.

Atmospheric turbulence degradation severely harms the task of face verification as the verification accuracy is reduced by more than 10$\%$ on average. There could be a significant drop (as much as $20\%$ for CFP$\_$FP) even though the face verification system is trained with \cite{guo2016ms}, which consists of over 5 million images. Also, as the task of restoration from turbulence is very challenging, the restoration results from other state-of-the-art method do not yield satisfactory results even they are trained with our dataset. Moreover, the proposed method restores the turbulence degraded images effectively even with a relatively small dataset.

\begin{figure}[t] 
\centering
\begin{subfigure}[t]{0.09\textwidth}
\includegraphics[width=\textwidth]{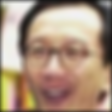}
\end{subfigure}
\begin{subfigure}[t]{0.09\textwidth}
\includegraphics[width=\textwidth]{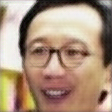}
\end{subfigure}
\begin{subfigure}[t]{0.09\textwidth}
\includegraphics[width=\textwidth]{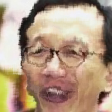}
\end{subfigure}
\begin{subfigure}[t]{0.09\textwidth}
\includegraphics[width=\textwidth]{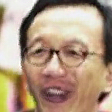}
\end{subfigure}
\begin{subfigure}[t]{0.09\textwidth}
\includegraphics[width=\textwidth]{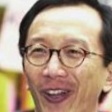}
\end{subfigure}
\\
\begin{subfigure}[t]{0.09\textwidth}
\includegraphics[width=\textwidth]{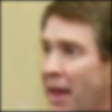}
\caption{}
\end{subfigure}
\begin{subfigure}[t]{0.09\textwidth}
\includegraphics[width=\textwidth]{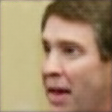}
\caption{}
\end{subfigure}
\begin{subfigure}[t]{0.09\textwidth}
\includegraphics[width=\textwidth]{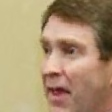}
\caption{}
\end{subfigure}
\begin{subfigure}[t]{0.09\textwidth}
\includegraphics[width=\textwidth]{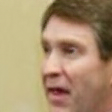}
\caption{}
\end{subfigure}
\begin{subfigure}[t]{0.09\textwidth}
\includegraphics[width=\textwidth]{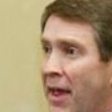}
\caption{}
\end{subfigure}
\caption{Visual performance comparison of the deblur function $G_d$ and deformation correction $G_b$ with the LFW dataset. (a) Blurry image. (b) Restored image of (a) by $G_d$. (c) Deformed image. (d) Restored image of (c) by $G_b$. (e) Groundtruth.}
\label{fig: performance of disentanglement}
\end{figure}

\begin{figure}[t] 
\centering
\begin{subfigure}[t]{0.23\textwidth}
\includegraphics[width=\textwidth]{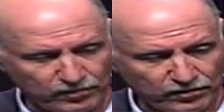}
\end{subfigure}
\begin{subfigure}[t]{0.23\textwidth}
\includegraphics[width=\textwidth]{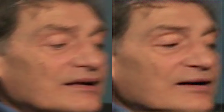}
\end{subfigure}
\\ \vspace{1mm}
\begin{subfigure}[t]{0.23\textwidth}
\includegraphics[width=\textwidth]{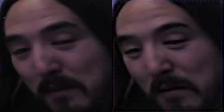}
\end{subfigure}
\begin{subfigure}[t]{0.23\textwidth}
\includegraphics[width=\textwidth]{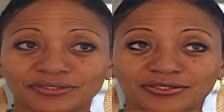}
\end{subfigure}

\caption{Visual performance comparison of original face image and the resultant image that passing through the pipeline with the original face image. Left: Original. Right: Ours.}
\label{fig: comparison}
\end{figure}

\begin{figure}[t] 
\centering
\begin{subfigure}[t]{0.11\textwidth}
\includegraphics[width=\textwidth]{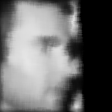}
\end{subfigure}
\begin{subfigure}[t]{0.11\textwidth}
\includegraphics[width=\textwidth]{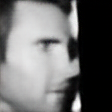}
\end{subfigure}
\begin{subfigure}[t]{0.11\textwidth}
\includegraphics[width=\textwidth]{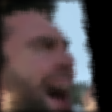}
\end{subfigure}
\begin{subfigure}[t]{0.11\textwidth}
\includegraphics[width=\textwidth]{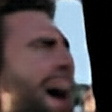}
\end{subfigure}
\\
\begin{subfigure}[t]{0.11\textwidth}
\includegraphics[width=\textwidth]{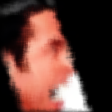}
\end{subfigure}
\begin{subfigure}[t]{0.11\textwidth}
\includegraphics[width=\textwidth]{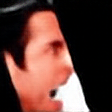}
\end{subfigure}
\begin{subfigure}[t]{0.11\textwidth}
\includegraphics[width=\textwidth]{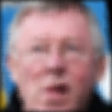}
\end{subfigure}
\begin{subfigure}[t]{0.11\textwidth}
\includegraphics[width=\textwidth]{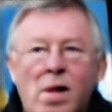}
\end{subfigure}

\caption{Failure cases. Left: AT. Right: Restored.}
\label{fig: fail}
\end{figure}

\subsection{Performance of the disentangled representation}
We try to disentangle the blur and deformation from atmospheric turbulence by training the deblur function $G_d$ and the deformation correction function $G_b$ with a commutative constraint. To see the performance of the disentanglement, $G_b$ and $G_d$ are tested. We try to use $G_d$ to deblur the blurry image and $G_b$ to correct the deformed images. Note that during the training, $G_b$ is only fed with the distorted image $\Tilde{I}$ and the deformation corrected image $G_d(\Tilde{I})$ of the distorted image.

We test $G_b$ and $G_d$ with $I_d$ and $I_b$ respectively, where $I$ are from the LFW dataset. The PSNR, SSIM and the accuracy of face verification are presented in Table \ref{tab:performance of G_b and G_d}. The visual performance is shown in Fig. \ref{fig: performance of disentanglement}. For the first row, the image is moderately blurred (Fig. \ref{fig: performance of disentanglement} (a)) and severely deformed (Fig. \ref{fig: performance of disentanglement} (c)). From Figs. \ref{fig: performance of disentanglement} (b) and (d), we see that $G_d$ and $G_b$ successfully remove the blur and deformation from the image and preserve the features of the subject. On the other hand, note that the image in the second row is a profile face with moderate blur (Fig. \ref{fig: performance of disentanglement} (a)) and mild distortion (Fig. \ref{fig: performance of disentanglement} (c)). Still, $G_d$ and $G_b$ successfully restore the degraded images. Moreover, the PSNR, SSIM and face verification results confirm that $G_d$ and $G_b$ restore the images, preserve shape and semantic information and are robust to severity of blur, deformation and pose.

\subsection{Robustness to clean images input}

We tested the case of feeding the groundtruth clean image into the proposed method. We tested $5222$ images from the IJB-C dataset \cite{maze2018iarpa}. The PSNR and SSIM of the resultant images after passing the original clean image to the pipeline are 33.89 and 0.980 respectively. The visual comparison result is shown in Fig. \ref{fig: comparison}. Both qualitative and quantitative results show that the proposed method is robust to non-blurred images. 

\subsection{Failure cases}
We have shown some failure cases in Fig. \ref{fig: fail}. The proposed method produces over-smoothed images if the turbulence degradation is very strong.
\begin{table}[t!]
\centering

\caption{PSNR, SSIM and face verification results for LFW dataset with $G_b$ and $G_d$.}
\label{tab:performance of G_b and G_d}
\begin{tabular}{|l || c | c | c | c|}
\hline
  
& $I_b$ & $I_d$ & $G_d(I_b)$ & $G_b(I_d)$   \\ \hline
PSNR & 25.33 & 29.78 & \textbf{28.72} & \textbf{29.93}  \\ \hline
SSIM & 0.895 & 0.958 & \textbf{0.931} & \textbf{0.961}  \\ \hline
Accuracy & 0.793 & 0.649 & \textbf{0.817} & \textbf{0.809} \\ \hline

\end{tabular}
\end{table}

\section{Conclusion}
In this paper, we proposed a single frame image restoration method ATFaceGAN, which is a generative algorithm to disentangle the turbulence distortion into blur and deformation and restores a sharp image. In order to disentangle turbulence, a deblur generator and a deformation correction generator are introduced. To further separate the blur and deformation, two paths are employed to produce two restored images. Finally, a fusion function combines the two restored images and generates one clean image. Ablation studies on each component demonstrate the effectiveness of different components. We have conducted extensive experiments on face restoration and face verification using the restored face images. Both quantitative and visual results show promising performance.

\section*{Acknowledgment}
This research is based upon work supported by the Office of the Director of National Intelligence (ODNI), Intelligence Advanced Research Projects Activity (IARPA), via IARPA R\&D Contract No. 2019-022600002. The views and conclusions contained herein are those of the authors and should not be interpreted as necessarily representing the official policies or endorsements, either expressed or implied, of the ODNI, IARPA, or the U.S. Government. The
U.S. Government is authorized to reproduce and distribute reprints for Governmental purposes notwithstanding any copyright annotation thereon.

\bibliographystyle{IEEEtran}
\bibliography{IEEEabrv,FG_2020/ref.bib}

\begin{thebibliography}{10}
\providecommand{\url}[1]{#1}
\csname url@samestyle\endcsname
\providecommand{\newblock}{\relax}
\providecommand{\bibinfo}[2]{#2}
\providecommand{\BIBentrySTDinterwordspacing}{\spaceskip=0pt\relax}
\providecommand{\BIBentryALTinterwordstretchfactor}{4}
\providecommand{\BIBentryALTinterwordspacing}{\spaceskip=\fontdimen2\font plus
\BIBentryALTinterwordstretchfactor\fontdimen3\font minus
  \fontdimen4\font\relax}
\providecommand{\BIBforeignlanguage}[2]{{%
\expandafter\ifx\csname l@#1\endcsname\relax
\typeout{** WARNING: IEEEtran.bst: No hyphenation pattern has been}%
\typeout{** loaded for the language `#1'. Using the pattern for}%
\typeout{** the default language instead.}%
\else
\language=\csname l@#1\endcsname
\fi
#2}}
\providecommand{\BIBdecl}{\relax}
\BIBdecl

\bibitem{hufnagel1964modulation}
R.~Hufnagel and N.~Stanley, ``Modulation transfer function associated with
  image transmission through turbulent media,'' \emph{JOSA}, vol.~54, no.~1,
  pp. 52--61, 1964.

\bibitem{roggemann1996imaging}
M.~C. Roggemann, B.~M. Welsh, and B.~R. Hunt, \emph{Imaging through
  turbulence}.\hskip 1em plus 0.5em minus 0.4em\relax CRC press, 1996.

\bibitem{oreifej2013simultaneous}
O.~Oreifej, X.~Li, and M.~Shah, ``Simultaneous video stabilization and moving
  object detection in turbulence,'' \emph{IEEE transactions on pattern analysis
  and machine intelligence}, vol.~35, no.~2, pp. 450--462, 2013.

\bibitem{chen2014detecting}
E.~Chen, O.~Haik, and Y.~Yitzhaky, ``Detecting and tracking moving objects in
  long-distance imaging through turbulent medium,'' \emph{Applied optics},
  vol.~53, no.~6, pp. 1181--1190, 2014.

\bibitem{pearson1976atmospheric}
J.~E. Pearson, ``Atmospheric turbulence compensation using coherent optical
  adaptive techniques,'' \emph{Applied optics}, vol.~15, no.~3, pp. 622--631,
  1976.

\bibitem{tyson2015principles}
R.~K. Tyson, \emph{Principles of adaptive optics}.\hskip 1em plus 0.5em minus
  0.4em\relax CRC press, 2015.

\bibitem{furhad2016restoring}
M.~H. Furhad, M.~Tahtali, and A.~Lambert, ``Restoring
  atmospheric-turbulence-degraded images,'' \emph{Applied optics}, vol.~55,
  no.~19, pp. 5082--5090, 2016.

\bibitem{hirsch2010efficient}
M.~Hirsch, S.~Sra, B.~Sch{\"o}lkopf, and S.~Harmeling, ``Efficient filter flow
  for space-variant multiframe blind deconvolution,'' in \emph{Computer Vision
  and Pattern Recognition (CVPR), 2010 IEEE Conference on}.\hskip 1em plus
  0.5em minus 0.4em\relax IEEE, 2010, pp. 607--614.

\bibitem{zhu2013removing}
X.~Zhu and P.~Milanfar, ``Removing atmospheric turbulence via space-invariant
  deconvolution,'' \emph{IEEE transactions on pattern analysis and machine
  intelligence}, vol.~35, no.~1, pp. 157--170, 2013.

\bibitem{micheli2014linear}
M.~Micheli, Y.~Lou, S.~Soatto, and A.~L. Bertozzi, ``A linear systems approach
  to imaging through turbulence,'' \emph{Journal of mathematical imaging and
  vision}, vol.~48, no.~1, pp. 185--201, 2014.

\bibitem{meinhardt2014implementation}
E.~Meinhardt-Llopis and M.~Micheli, ``Implementation of the centroid method for
  the correction of turbulence,'' \emph{Image Processing On Line}, vol.~4, pp.
  187--195, 2014.

\bibitem{lou2013video}
Y.~Lou, S.~H. Kang, S.~Soatto, and A.~L. Bertozzi, ``Video stabilization of
  atmospheric turbulence distortion,'' \emph{Inverse Problems \& Imaging},
  vol.~7, no.~3, 2013.

\bibitem{lau2019variational}
C.~P. Lau, Y.~H. Lai, and L.~M. Lui, ``Variational models for joint subsampling
  and reconstruction of turbulence-degraded images,'' \emph{Journal of
  Scientific Computing}, vol.~78, no.~3, pp. 1488--1525, 2019.

\bibitem{lau2019restoration}
C.~P. Lau, Y.~H. Lai, and R.~L.~M. Lui, ``Restoration of atmospheric
  turbulence-distorted images via rpca and quasiconformal maps,'' \emph{Inverse
  Problems}, 2019.

\bibitem{chak2018subsampled}
W.~H. Chak, C.~P. Lau, and L.~M. Lui, ``Subsampled turbulence removal
  network,'' \emph{arXiv preprint arXiv:1807.04418}, 2018.

\bibitem{Chrysos_2017_CVPR_Workshops}
G.~G. Chrysos and S.~Zafeiriou, ``Deep face deblurring,'' in \emph{The IEEE
  Conference on Computer Vision and Pattern Recognition (CVPR) Workshops}, July
  2017.

\bibitem{shen2018deep}
Z.~Shen, W.-S. Lai, T.~Xu, J.~Kautz, and M.-H. Yang, ``Deep semantic face
  deblurring,'' in \emph{Proceedings of the IEEE Conference on Computer Vision
  and Pattern Recognition}, 2018, pp. 8260--8269.

\bibitem{lu2019unsupervised}
B.~Lu, J.-C. Chen, and R.~Chellappa, ``Unsupervised domain-specific deblurring
  via disentangled representations,'' \emph{arXiv preprint arXiv:1903.01594},
  2019.

\bibitem{chen2018fsrnet}
Y.~Chen, Y.~Tai, X.~Liu, C.~Shen, and J.~Yang, ``Fsrnet: End-to-end learning
  face super-resolution with facial priors,'' in \emph{Proceedings of the IEEE
  Conference on Computer Vision and Pattern Recognition}, 2018, pp. 2492--2501.

\bibitem{Yu_2018_CVPR}
X.~Yu, B.~Fernando, R.~Hartley, and F.~Porikli, ``Super-resolving very
  low-resolution face images with supplementary attributes,'' in \emph{The IEEE
  Conference on Computer Vision and Pattern Recognition (CVPR)}, June 2018.

\bibitem{Yu_2018_ECCV}
X.~Yu, B.~Fernando, B.~Ghanem, F.~Porikli, and R.~Hartley, ``Face
  super-resolution guided by facial component heatmaps,'' in \emph{The European
  Conference on Computer Vision (ECCV)}, September 2018.

\bibitem{arjovsky2017wasserstein}
M.~Arjovsky, S.~Chintala, and L.~Bottou, ``Wasserstein gan,'' \emph{arXiv
  preprint arXiv:1701.07875}, 2017.

\bibitem{bansal2017umdfaces}
A.~Bansal, A.~Nanduri, C.~D. Castillo, R.~Ranjan, and R.~Chellappa, ``Umdfaces:
  An annotated face dataset for training deep networks,'' in \emph{2017 IEEE
  International Joint Conference on Biometrics (IJCB)}.\hskip 1em plus 0.5em
  minus 0.4em\relax IEEE, 2017, pp. 464--473.

\bibitem{aubailly2009automated}
M.~Aubailly, M.~A. Vorontsov, G.~W. Carhart, and M.~T. Valley, ``Automated
  video enhancement from a stream of atmospherically-distorted images: the
  lucky-region fusion approach,'' in \emph{Proc. SPIE}, vol. 7463, 2009, p.
  74630C.

\bibitem{vorontsov2001anisoplanatic}
M.~A. Vorontsov and G.~W. Carhart, ``Anisoplanatic imaging through turbulent
  media: image recovery by local information fusion from a set of
  short-exposure images,'' \emph{JOSA A}, vol.~18, no.~6, pp. 1312--1324, 2001.

\bibitem{xie2016removing}
Y.~Xie, W.~Zhang, D.~Tao, W.~Hu, Y.~Qu, and H.~Wang, ``Removing turbulence
  effect via hybrid total variation and deformation-guided kernel regression,''
  \emph{IEEE Transactions on Image Processing}, vol.~25, no.~10, pp.
  4943--4958, 2016.

\bibitem{radford2015unsupervised}
A.~Radford, L.~Metz, and S.~Chintala, ``Unsupervised representation learning
  with deep convolutional generative adversarial networks,'' \emph{arXiv
  preprint arXiv:1511.06434}, 2015.

\bibitem{kupyn2018deblurgan}
O.~Kupyn, V.~Budzan, M.~Mykhailych, D.~Mishkin, and J.~Matas, ``Deblurgan:
  Blind motion deblurring using conditional adversarial networks,'' in
  \emph{Proceedings of the IEEE Conference on Computer Vision and Pattern
  Recognition}, 2018, pp. 8183--8192.

\bibitem{he2016deep}
K.~He, X.~Zhang, S.~Ren, and J.~Sun, ``Deep residual learning for image
  recognition,'' in \emph{Proceedings of the IEEE conference on computer vision
  and pattern recognition}, 2016, pp. 770--778.

\bibitem{ulyanov2016instance}
D.~Ulyanov, A.~Vedaldi, and V.~Lempitsky, ``Instance normalization: The missing
  ingredient for fast stylization,'' \emph{arXiv preprint arXiv:1607.08022},
  2016.

\bibitem{nair2010rectified}
V.~Nair and G.~E. Hinton, ``Rectified linear units improve restricted boltzmann
  machines,'' in \emph{Proceedings of the 27th international conference on
  machine learning (ICML-10)}, 2010, pp. 807--814.

\bibitem{gulrajani2017improved}
I.~Gulrajani, F.~Ahmed, M.~Arjovsky, V.~Dumoulin, and A.~C. Courville,
  ``Improved training of wasserstein gans,'' in \emph{Advances in Neural
  Information Processing Systems}, 2017, pp. 5767--5777.

\bibitem{isola2017image}
P.~Isola, J.-Y. Zhu, T.~Zhou, and A.~A. Efros, ``Image-to-image translation
  with conditional adversarial networks,'' in \emph{Proceedings of the IEEE
  conference on computer vision and pattern recognition}, 2017, pp. 1125--1134.

\bibitem{li2016precomputed}
C.~Li and M.~Wand, ``Precomputed real-time texture synthesis with markovian
  generative adversarial networks,'' in \emph{European Conference on Computer
  Vision}.\hskip 1em plus 0.5em minus 0.4em\relax Springer, 2016, pp. 702--716.

\bibitem{xu2015empirical}
B.~Xu, N.~Wang, T.~Chen, and M.~Li, ``Empirical evaluation of rectified
  activations in convolutional network,'' \emph{arXiv preprint
  arXiv:1505.00853}, 2015.

\bibitem{simonyan2014very}
K.~Simonyan and A.~Zisserman, ``Very deep convolutional networks for
  large-scale image recognition,'' \emph{arXiv preprint arXiv:1409.1556}, 2014.

\bibitem{deng2009imagenet}
J.~Deng, W.~Dong, R.~Socher, L.-J. Li, K.~Li, and L.~Fei-Fei, ``Imagenet: A
  large-scale hierarchical image database,'' in \emph{2009 IEEE conference on
  computer vision and pattern recognition}.\hskip 1em plus 0.5em minus
  0.4em\relax Ieee, 2009, pp. 248--255.

\bibitem{huang2008labeled}
G.~B. Huang, M.~Mattar, T.~Berg, and E.~Learned-Miller, ``Labeled faces in the
  wild: A database forstudying face recognition in unconstrained
  environments,'' in \emph{Workshop on faces in'Real-Life'Images: detection,
  alignment, and recognition}, 2008.

\bibitem{sengupta2016frontal}
S.~Sengupta, J.-C. Chen, C.~Castillo, V.~M. Patel, R.~Chellappa, and D.~W.
  Jacobs, ``Frontal to profile face verification in the wild,'' in \emph{2016
  IEEE Winter Conference on Applications of Computer Vision (WACV)}.\hskip 1em
  plus 0.5em minus 0.4em\relax IEEE, 2016, pp. 1--9.

\bibitem{moschoglou2017agedb}
S.~Moschoglou, A.~Papaioannou, C.~Sagonas, J.~Deng, I.~Kotsia, and
  S.~Zafeiriou, ``Agedb: the first manually collected, in-the-wild age
  database,'' in \emph{Proceedings of the IEEE Conference on Computer Vision
  and Pattern Recognition Workshops}, 2017, pp. 51--59.

\bibitem{zheng2017cross}
T.~Zheng, W.~Deng, and J.~Hu, ``Cross-age lfw: A database for studying
  cross-age face recognition in unconstrained environments,'' \emph{arXiv
  preprint arXiv:1708.08197}, 2017.

\bibitem{zheng2018cross}
T.~Zheng and W.~Deng, ``Cross-pose lfw: A database for studying crosspose face
  recognition in unconstrained environments,'' \emph{Beijing University of
  Posts and Telecommunications, Tech. Rep}, pp. 18--01, 2018.

\bibitem{cao2018vggface2}
Q.~Cao, L.~Shen, W.~Xie, O.~M. Parkhi, and A.~Zisserman, ``Vggface2: A dataset
  for recognising faces across pose and age,'' in \emph{2018 13th IEEE
  International Conference on Automatic Face \& Gesture Recognition (FG
  2018)}.\hskip 1em plus 0.5em minus 0.4em\relax IEEE, 2018, pp. 67--74.

\bibitem{paszke2017automatic}
A.~Paszke, S.~Gross, S.~Chintala, G.~Chanan, E.~Yang, Z.~DeVito, Z.~Lin,
  A.~Desmaison, L.~Antiga, and A.~Lerer, ``Automatic differentiation in
  pytorch,'' 2017.

\bibitem{kingma2014adam}
D.~P. Kingma and J.~Ba, ``Adam: A method for stochastic optimization,''
  \emph{arXiv preprint arXiv:1412.6980}, 2014.

\bibitem{xu2017high}
Y.~Xu, Y.~Cheng, J.~Zhao, Z.~Wang, L.~Xiong, K.~Jayashree, H.~Tamura,
  T.~Kagaya, S.~Shen, S.~Pranata \emph{et~al.}, ``High performance large scale
  face recognition with multi-cognition softmax and feature retrieval,'' in
  \emph{Proceedings of the IEEE International Conference on Computer Vision},
  2017, pp. 1898--1906.

\bibitem{guo2016ms}
Y.~Guo, L.~Zhang, Y.~Hu, X.~He, and J.~Gao, ``Ms-celeb-1m: A dataset and
  benchmark for large-scale face recognition,'' in \emph{European Conference on
  Computer Vision}.\hskip 1em plus 0.5em minus 0.4em\relax Springer, 2016, pp.
  87--102.

\bibitem{maze2018iarpa}
B.~Maze, J.~Adams, J.~A. Duncan, N.~Kalka, T.~Miller, C.~Otto, A.~K. Jain,
  W.~T. Niggel, J.~Anderson, J.~Cheney \emph{et~al.}, ``Iarpa janus
  benchmark-c: Face dataset and protocol,'' in \emph{2018 International
  Conference on Biometrics (ICB)}.\hskip 1em plus 0.5em minus 0.4em\relax IEEE,
  2018, pp. 158--165.

\end{thebibliography}



%



\end{document}